\ifdefined\pdfminorversion
\fi

\PassOptionsToPackage{table}{xcolor}
\PassOptionsToPackage{breaklinks=true,colorlinks,citecolor=blue,urlcolor=blue,linkcolor=blue,bookmarks=false}{hyperref}
\PassOptionsToPackage{noabbrev,nameinlink,capitalize}{cleveref}

\documentclass[numbers]{april_aigc}

\microtypesetup{expansion=false}

\usepackage{cmap}
\usepackage[utf8]{inputenc}
\usepackage[T1]{fontenc}
\usepackage{float}
\usepackage{url}
\usepackage{graphicx}
\usepackage[percent]{overpic}
\usepackage{booktabs}
\usepackage{amsmath}
\usepackage{amssymb}
\usepackage{amsfonts}
\usepackage{microtype}
\usepackage{xspace}
\usepackage{placeins}
\usepackage{wrapfig}
\usepackage{tikz}
\usepackage{multicol}
\usepackage{multirow}
\usepackage{makecell}
\usepackage{tabularx}
\usepackage{adjustbox}
\usepackage{colortbl}
\usepackage{pifont}
\usepackage{enumitem}
\usepackage[normalem]{ulem}
\usepackage{algorithm}
\usepackage{algorithmic}
\usepackage{listings}
\usepackage{etoolbox}
\usepackage{caption}

\IfFileExists{glyphtounicode.tex}{\input{glyphtounicode}}{}
\ifdefined\pdfgentounicode
\fi
\DisableLigatures{encoding = *, family = *}

\graphicspath{{fig/}}

\definecolor{linkcolor}{named}{aprilblue}
\definecolor{urlcolor}{RGB}{255,105,180}
\definecolor{citecolor}{RGB}{66,168,235}
\definecolor{lightgray}{rgb}{0.8, 0.8, 0.8}
\definecolor{darkgreen}{rgb}{0.00, 0.81, 0.78}
\definecolor{gray_tab}{RGB}{220, 220, 220}
\definecolor{blue_tab}{RGB}{227, 240, 251}
\definecolor{oran_tab}{RGB}{252, 242, 237}
\definecolor{whit_tab}{RGB}{255, 255, 255}
\definecolor{green_code}{RGB}{55, 126, 34}
\definecolor{oursrowblue}{RGB}{232,242,255}
\definecolor{boxteal}{RGB}{70,160,160}

\newcommand{\method}{ChatImage}

\renewcommand\authorformat[2][]{{\sffamily \bfseries #2\if\relax#1\relax\else\textsuperscript{#1}\fi}}

\newcommand{\best}[1]{\textbf{#1}}

\newcommand{\na}{-{}-}

\makeatletter
\AfterEndEnvironment{algorithm}{\let\@algcomment\relax}
\AtEndEnvironment{algorithm}{\kern2pt\hrule\relax\vskip3pt\@algcomment}
\let\@algcomment\relax
\newcommand\algcomment[1]{\def\@algcomment{\footnotesize#1}}
\renewcommand\fs@ruled{\def\@fs@cfont{\bfseries}\let\@fs@capt\floatc@ruled
  \def\@fs@pre{\hrule height.8pt depth0pt \kern2pt}%
  \def\@fs@post{}%
  \def\@fs@mid{\kern2pt\hrule\kern2pt}%
  \let\@fs@iftopcapt\iftrue}
\makeatother

\DeclareCaptionFormat{custom}{{\color{aprilblue}\sffamily\textbf{#1 #2}} #3}
\titleformat*{\section}{\color{aprilblue}\Large\sffamily\bfseries}
\titleformat*{\subsection}{\color{aprilblue}\large\sffamily\bfseries}
\titleformat*{\subsubsection}{\color{aprilblue}\normalsize\sffamily\bfseries}

\usepackage{fancyhdr}
\newif\ifshowlogo
\showlogotrue
\newcommand{\insertlogo}{%
  \ifshowlogo
    \IfFileExists{assets/april_logo1.png}%
    {\includegraphics[height=0.68cm]{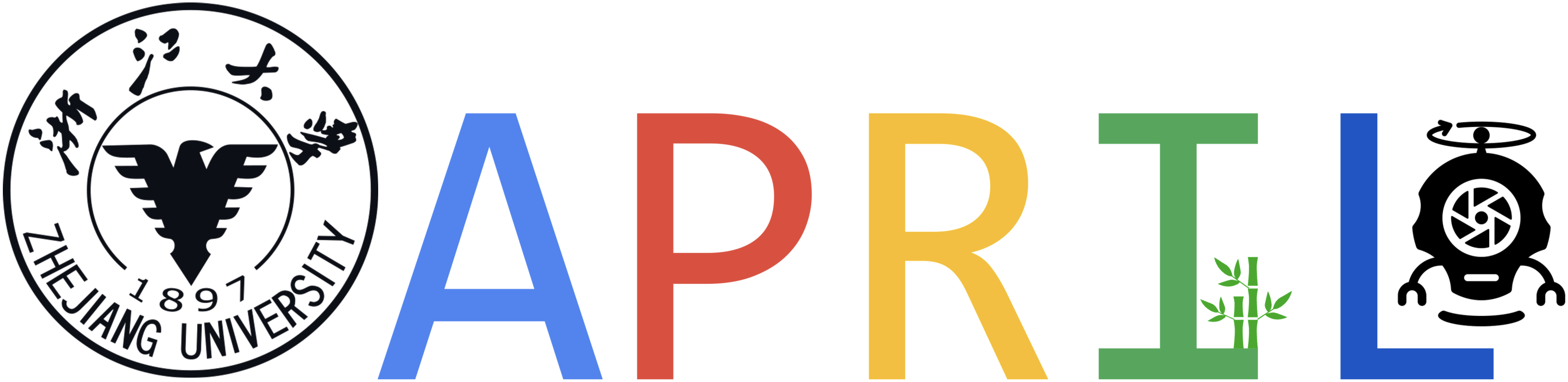}}%
    {}%
  \fi
}
\newif\ifshowtoc
\showtocfalse

\hypersetup{
  colorlinks=true,
  linkcolor=linkcolor,
  urlcolor=urlcolor,
  citecolor=citecolor,
  pdfauthor={Wencan Jiang, Jiangning Zhang, Yong Liu},
  pdftitle={ChatImage: Navigating Long-Form LLM Answers through Interactive Images}
}

\renewcommand{\title}[1]{\def\titlelist{{\fontsize{20pt}{28pt}\selectfont\sffamily\bfseries #1}}}
\title{ChatImage: Navigating Long-Form LLM Answers through Interactive Images}

\author[1]{Wencan Jiang}
\author[1,\raisebox{-0.2em}{\includegraphics[height=0.85em]{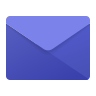}}]{Jiangning Zhang}
\author[1]{Yong Liu}

\affiliation[1]{Zhejiang University}
\abstract{
Large Language Models (LLMs) can produce detailed answers to complex queries,
but these answers are typically presented as dense linear text, which makes
fine-grained inspection, navigation, and return visits difficult. We present
\method{}, a system that converts long-form LLM answers into
\textbf{interactive visual images}. Given a textual answer, \method{} first
normalizes its content into \textbf{structured visual modules}, plans a visual
layout, and renders a coherent image. It then applies a \textbf{second
grounding pass} to the rendered image with vision grounding models such as
LocateAnything and MiMo-Vision, with optional SAM-style mask refinement, to
identify the visible regions that should support interaction. From these
grounded regions, \method{} overlays \textbf{transparent clickable hotspots}
on the image. Each hotspot opens a detail panel and a region-scoped follow-up
thread, allowing the user to inspect and query a specific part of the answer
without re-reading the full response. Instead of treating planned coordinates
as the final interaction geometry, \method{} uses them as priors and
\textbf{grounds the interaction targets after rendering}, which improves
consistency between visual content and clickable regions. We release a
reference implementation and introduce a \textbf{30-question benchmark}
covering infographic, map, and scene-based answer formats. Evaluation with
configured external models reports interaction-loop completion, a strict
visual-alignment gate, and a SAM-based mask-completeness diagnostic.
}

\coverdate{\today}
\covercorrespondence{186368@zju.edu.cn}
\coversourcecode{\href{https://github.com/wencanjiang/ChatImage}{github.com/wencanjiang/ChatImage}}
\coverproject{\href{https://wencanjiang.github.io/ChatImage/}{wencanjiang.github.io/ChatImage}}

\begin{document}

\maketitle
\thispagestyle{plain}

\ifshowtoc
  \clearpage
  \setcounter{tocdepth}{2}
  \tableofcontents
  \vspace{1cm}
  \clearpage
\fi

\section{Introduction}
\label{sec:introduction}

Large Language Models (LLMs) have become a common interface for answering
complex questions~\citep{openai2023gpt4,anthropic2024claude}. Their answers,
however, are still delivered mainly as extended prose. This format is effective
for short factual queries, but it becomes less suitable when an answer contains
\textbf{internal structure}, such as multiple entities, comparisons, procedural
stages, or spatial relations. In such cases, readers often need to locate a
specific module, inspect its context, and return to it later. Plain text can
encode this information, but it provides \textbf{limited support for
navigation}.

Recent progress in text-to-image generation~\citep{ramesh2022dalle2,saharia2022imagen,podell2024sdxl}
and vision-language understanding~\citep{liu2023llava,chen2024internvl}
suggests a complementary presentation format: \textbf{rendering the answer as
an image}. A generated infographic, map, or scene can expose modules,
hierarchies, routes, and spatial relationships in a single view. Static images,
however, lose the explanatory structure associated with each region and provide
no mechanism for region-level follow-up. A direct overlay of planned boxes is
also unreliable, because image generators do not consistently place content at
the requested coordinates~\citep{feng2023layoutgpt,cheng2024intelligent}. If
the interaction layer is copied from the pre-generation layout, a user may
click an empty region or an unintended visual element.

\begin{figure}[t!]
    \centering
    \includegraphics[width=1.0\textwidth]{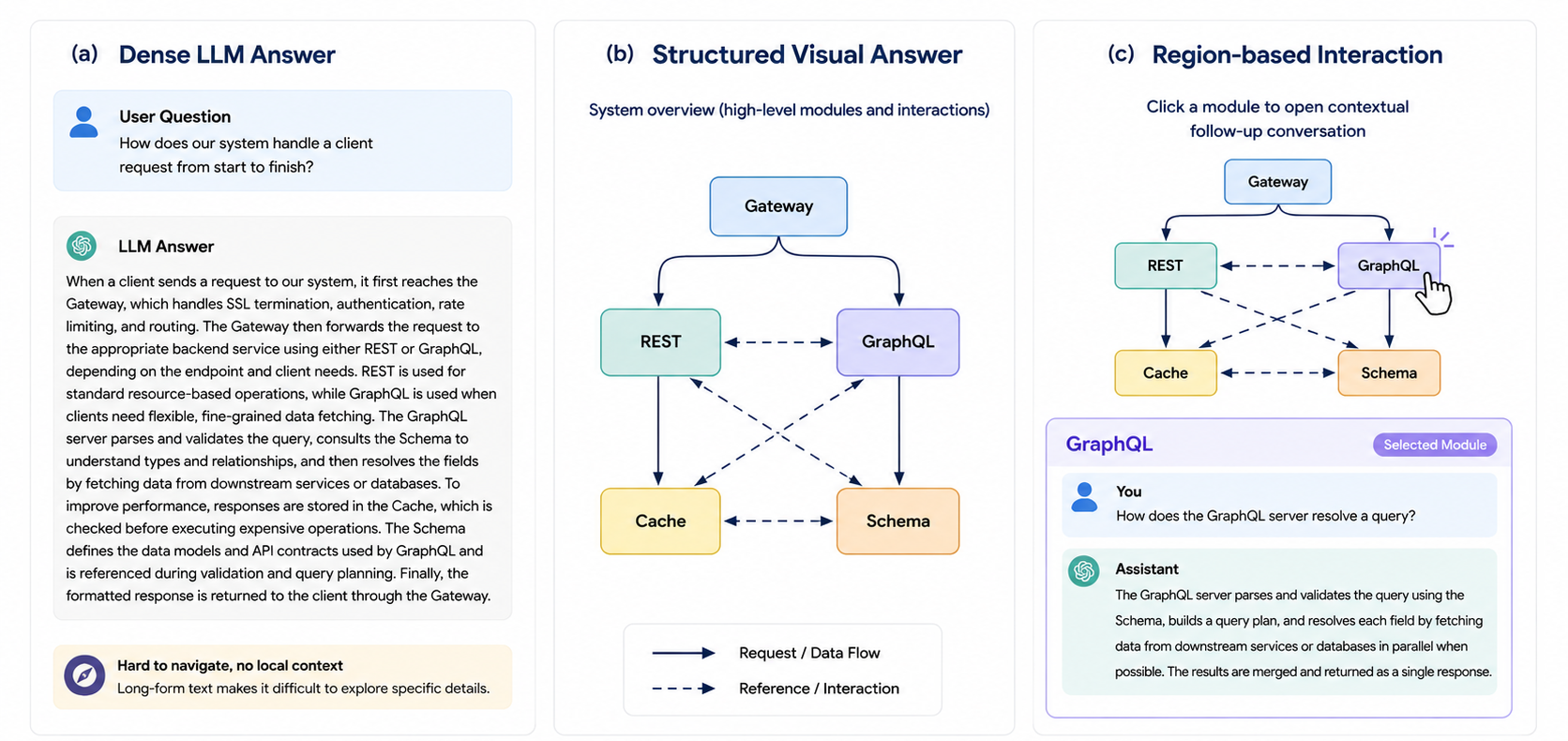}
    \caption{\textbf{\method{} turns a long-form LLM answer into an
    interactive visual image.} Given a user question, \method{} generates an
    image whose regions are clickable hotspots. Each hotspot opens a detail
    panel with the region's explanation and an in-context follow-up thread,
    enabling targeted inspection of individual answer regions.}
    \label{fig:teaser}
    \vspace{-0.6em}
\end{figure}

We propose \method{}, a system that converts a long-form LLM answer into an
\textbf{interactive visual image}: a generated image with transparent hotspots
placed over visible answer regions (Figure~\ref{fig:teaser}). Each hotspot
opens the explanation associated with that region and maintains a local
follow-up thread. The central technical problem is \textbf{alignment}: a
hotspot should cover the object, panel, landmark, or diagram region visible to
the user, rather than the box specified before image generation.

\method{} addresses this problem through a \textbf{generate-then-ground
interaction loop}. The system converts the answer into a \textbf{small set of
visual modules}, plans an approximate layout, synthesizes an image prompt, and
generates the image. A \textbf{second visual grounding pass} then localizes the
rendered modules and places \textbf{clickable hotspots} on the discovered
regions. This design separates \textbf{what the answer should contain} from
\textbf{where the generator renders it}. The resulting artifact supports
\textbf{region-level exploration}: the user selects a visible region, inspects
the corresponding detail and local preview, and asks follow-up questions
without restating the full context.
The implementation uses a provider-agnostic backend for text, image, and vision
models, with deterministic local models for reproducible testing and configured
external models for full-generation evaluation.

To our knowledge, prior work has not studied LLM answers as navigable,
region-interactive visual images with \textbf{vision-grounded hotspots}. The
main contributions are:
\begin{itemize}
    \item \textbf{Interactive visual answers.} We formulate the task of
    presenting a long-form LLM answer as an explorable image with clickable
    hotspots, detail panels, and per-region follow-up threads.
    \item \textbf{Generate-then-ground alignment.} We decouple structured
    content from visual placement: the system first renders the image, then
    uses visual grounding and optional SAM-family mask refinement to place
    hotspots on visible regions rather than on a fixed layout grid.
    \item \textbf{Reference implementation.} We provide a provider-agnostic
    implementation with deterministic local execution, configured external
    model execution, local persistence, per-hotspot follow-up threads, and
    calibration tools for inspecting hotspot bounds.
    \item \textbf{Benchmark and grounding evaluation.} We introduce a
    30-question benchmark spanning multiple answer formats and evaluate the
    full interaction loop with completion checks and a strict visual-alignment
    gate.
\end{itemize}

\section{Related Work}
\label{sec:related_work}

\method{} is related to text-to-image generation, visual grounding,
interactive visualization, and structured LLM output. The specific problem
addressed here, converting an LLM answer into a region-interactive visual image,
has received limited direct attention. We therefore discuss the closest
technical areas and identify the gap addressed by \method{}.

\subsection{Text-to-Image and Controllable Generation}
Diffusion models~\citep{ramesh2022dalle2,saharia2022imagen,rombach2022ldm,podell2024sdxl}
can produce high-fidelity images, but they do not reliably satisfy precise
spatial instructions. A prompt that requests a module at a specified canvas
location does not guarantee exact placement~\citep{feng2023layoutgpt,cheng2024intelligent}.
Controllable methods such as ControlNet~\citep{zhang2023controlnet},
GLIGEN~\citep{li2023gligen}, training-free layout-to-image~\citep{chen2024trainingfree},
and InfographicNLP~\citep{lu2023infographicnlp} impose additional structure,
but their primary objective is image synthesis rather than the construction of
interactive regions aligned with a structured answer. \method{} uses
text-to-image generation as a rendering backend and estimates region locations
after generation through visual grounding.

\subsection{Visual Grounding and Segmentation}
\begin{sloppypar}
Referring localization methods such as MDETR~\citep{kamath2021mdetr},
TransVG~\citep{deng2021transvg}, UNINEXT~\citep{yan2023uninext}, Grounding
DINO~\citep{liu2024groundingdino}, and YOLO-World~\citep{cheng2024yoloworld},
as well as
grounding-oriented VLMs~\citep{nvidia2025locateanything,xiaomi2025mimovl,liu2023llava,chen2024internvl}
return boxes for descriptive prompts, while SAM/SAM2~\citep{kirillov2023sam,ravi2024sam2}
can refine a box into a mask. Broader multimodal in-context learning work, such
as UniICL~\citep{xu2026uniicl}, studies how demonstrations condition unified
multimodal models across understanding and generation tasks. \method{} instead
uses multimodal models as grounding components inside an interactive artifact
construction loop. The goal is not to detect a predefined object category, but
to align structured answer regions with their rendered positions in a newly
generated image.
\end{sloppypar}

\subsection{Interactive Visualization}
Interactive image annotation~\citep{hochheiser2004interactive,munzner2014visualization}
and VQA~\citep{antol2015vqa,goyal2017vqav2,mathew2021docvqa,masry2022chartqa}
study fixed images rather than answer-generated images. VQA also treats the
image as input to be queried, whereas \method{} treats the image as the answer
interface itself. Structured-output work such as chain-of-thought~\citep{wei2022cot},
ReAct~\citep{yao2023react}, tool use~\citep{schick2023toolformer},
and RAG~\citep{lewis2020rag,gao2024ragsurvey} structures \emph{reasoning}, not
the output \emph{modality}; the answer stays text. Efficiency-oriented agent
and model-compression work, such as SPIKE's selective long-horizon controller
for game agents~\citep{jiang2026spike} and token-adaptive knowledge
distillation for LLMs~\citep{xie2026adakd}, addresses the cost of reasoning or
model serving, which is complementary to the interaction-focused problem studied
here. Mind-map and
chart-generation methods~\citep{zhang2014mindmap,hu2024llmmindmap,obeid2020charttotext,tang2023chartlama}
visualize structure but as static outputs without image-level region interaction
or grounding. \method{} combines these threads in an underexplored setting: an
LLM answer rendered as a navigable, region-interactive visual image with
vision-grounded hotspots.

\section{Method}
\label{sec:method}

\method{} converts a user's question into an interactive visual image by
separating content specification from spatial realization. The system first
determines \emph{what} the answer should contain, and only after image
generation estimates \emph{where} the corresponding content appears. This
section describes the end-to-end pipeline
(Section~\ref{sec:method_loop}), the structured answer normalization and
layout planning (Section~\ref{sec:method_structure}), the two-pass vision
alignment that grounds hotspots to real image content
(Section~\ref{sec:method_alignment}), and the interactive hotspot layer and
per-region follow-up threads (Section~\ref{sec:method_interaction}).

\subsection{End-to-End Interaction Loop}
\label{sec:method_loop}

Given a user question $q$, \method{} produces an interactive visual image
$\mathcal{I}$ with a set of clickable hotspots $\mathcal{H}$. The pipeline has
two passes. The first pass specifies the answer content and generates the
image; the second pass localizes the rendered content and derives the clickable
hotspot layer from the resulting evidence.

\paragraph{Pass 1: Structure, layout, and image generation.}
\begin{enumerate}[leftmargin=*,topsep=2pt,itemsep=1pt]
    \item \textbf{Raw answer.} Obtain a long-form LLM answer $a$ to $q$ from
    a text model.
    \item \textbf{Normalization.} Normalize $a$ into a structured visual
    specification $\mathcal{S}$, a set of modules $\{m_i\}$ and auxiliary
    modules, each with a title, detail text, region prompt, and metadata
    (Section~\ref{sec:method_structure}).
    \item \textbf{Layout planning.} Plan a layout $\mathcal{L}$ that assigns
    each module a region $r_i$ with normalized bounds in $[0,1]^4$.
    \item \textbf{Image prompt synthesis.} Derive an image prompt from
    $\mathcal{S}$ and $\mathcal{L}$, encoding the visual style and the
    intended region labels and positions.
    \item \textbf{Image generation.} Generate the image via a text-to-image
    model, obtaining $\mathcal{I}$ and its real pixel dimensions
    $(W, H)$ parsed from the image header (not the request, since generators
    do not guarantee exact sizes).
\end{enumerate}

\paragraph{Pass 2: Vision alignment and hotspot derivation.}
\begin{enumerate}[leftmargin=*,topsep=2pt,itemsep=1pt,resume]
    \item \textbf{Visual grounding.} Send $\mathcal{I}$, $(W,H)$, and the
    modules with their planned bounds to a visual grounding model, which
    returns a grounded box $b_i$ for each rendered module
    (Section~\ref{sec:method_alignment}).
    \item \textbf{Mask refinement (optional).} Refine each $b_i$, when enabled,
    with the SAM3 adapter to obtain a pixel mask for tighter previews.
    \item \textbf{Hotspot derivation.} Copy the grounded bounds into
    rectangular hotspot geometry and render transparent, clickable hotspots
    over $\mathcal{I}$ (Section~\ref{sec:method_interaction}).
\end{enumerate}

\method{} \textbf{does not assume the image generator follows the planned
layout}. The pipeline treats layout planning as a prior and visual grounding as
the source of final hotspot geometry. This design avoids requiring precise
spatial control from diffusion models, which current generators do not reliably
provide~\citep{feng2023layoutgpt}, and ties clickable regions to rendered
content instead of planned coordinates.

\subsection{Structured Answer Normalization and Layout}
\label{sec:method_structure}

\paragraph{Visual specification.}
The raw answer $a$ is parsed into a visual specification $\mathcal{S}$ with a
visual mode (infographic, map, scene, or poster), a summary, and a list of
modules. Each module $m_i$ carries:
\textit{title} (the hotspot label), \textit{detail} (the explanation shown in
the detail panel), \textit{imageText} (short on-image caption),
\textit{regionPrompt} (a locator phrase for visual grounding), and metadata
(\textit{regionKind}, \textit{iconHint}, \textit{maskPolicy}, locator queries).
Because LLMs may echo prompt scaffolding into user-facing text, a sanitization
pass removes visual-system vocabulary, meta-instructions, and obvious prompt
fragments before display. Short labels are used as fallbacks when a detail
field becomes unusable. If no valid structure is returned, the reference
implementation can instantiate deterministic specifications for common
infographic, map, and scene topics, which supports controlled testing of the
downstream interaction pipeline.

\paragraph{Layout planning.}
A layout planner assigns each module a region $r_i$ with normalized bounds
$\in [0,1]^4$ relative to the image canvas. Layouts are organized by
family, including grid, flow (swimlane), compare, hub, timeline, matrix, and
freeform layouts, selected by the visual mode and relation type. For maps, a
semantic-slot catalog assigns regions by keyword matching (e.g., "entrance"
maps to bottom-center, "core area" maps to center); for infographics, index-based
templates place modules in flow or grid order. These \emph{planned} bounds
serve as priors for the grounding step and as a fallback when grounding fails,
but they are \emph{not} assumed to match the generated image exactly.

\subsection{Two-Pass Vision Alignment}
\label{sec:method_alignment}

The central technical component is the alignment of structured regions to the
actual generated image. \method{} sends the image, its real dimensions, the
modules, and their planned bounds to a vision backend that runs a
grounding-and-repair loop:

\paragraph{Grounding model.}
The grounding step is model-agnostic: any vision backend that can return a
normalized bounding box for a locator phrase can be used. The reference
implementation supports a dedicated referring-localization model
(\textbf{LocateAnything-3B}~\citep{nvidia2025locateanything}) and a general
vision-language model used in structured grounding mode
(\textbf{MiMo-Vision}~\citep{xiaomi2025mimovl}). For each module the backend
receives the generated image plus locator phrases built from the module title,
region prompt, and semantic hints. It returns candidate boxes in normalized
coordinates. The system compares the candidates with the planned layout,
selects the most plausible box, and may re-ground within a crop when the
initial candidate is uncertain. This is a \emph{referring localization} step:
the system recovers the rendered location of an answer module rather than
detecting a predefined object class.

\paragraph{Fallback hierarchy.}
When the grounding backend rejects a module or is unavailable, \method{}
uses a graded fallback hierarchy: a general VLM-based grounding pass, local
OCR, and finally the planned bounds. Each module is tagged with its alignment
source (e.g., \textit{locateanything}, \textit{mimo-vision},
\textit{sam3-refined-planned}, \textit{planned-fallback},
\textit{vision-low-confidence}) so the system can report alignment quality
per region rather than as a single aggregate score.

\paragraph{Robustness to partial failure.}
A single missing or low-confidence grounding should not invalidate the entire
result. The system rejects groundings that resemble header strips or cross-panel
artifacts (falling back to planned bounds for those) while keeping
far-from-planned groundings that still localize the correct region (tagged
low-confidence). Modules entirely absent from the grounding response retain
their planned bounds and are flagged, rather than discarding the alignments of
all correctly-grounded modules. Each grounded box is repaired for
clickability by enforcing a minimum area and safe margin before becoming a
hotspot.

\paragraph{SAM3 mask refinement (optional).}
For tighter visual previews, an interactive box-prompted segmentation model in
the SAM family~\citep{kirillov2023sam,ravi2024sam2} takes each grounded box as
a prompt and predicts a pixel mask, from which a tightened box and polygon are
derived. The mask is used for preview rendering, including cutouts, organic
feathering, and soft fades, while click hit-testing remains rectangular in the normal
interaction loop. This design preserves stable interaction behavior even when a
mask is visually useful but not sufficiently reliable to define the clickable
shape.

\subsection{Interactive Hotspot Layer and Follow-up Threads}
\label{sec:method_interaction}

\paragraph{Hotspot rendering.}
Each hotspot is a transparent, absolutely-positioned button over the image at
its grounded coordinates. In normal use, hotspots are fully invisible so the
generated image remains visually unobstructed; a hover state provides a soft
visual cue without altering the image. A calibration mode toggles visible
outlines and per-hotspot labels for inspecting and correcting bounds.

\paragraph{Detail panels and previews.}
Clicking a hotspot opens a centered detail panel showing the module's title,
sanitized detail text, a localized preview of the image region, and a
follow-up input. The preview variant is chosen by a strategy module based on
the hotspot shape and mask availability: cutout (checkerboard background),
organic feathering, soft radial fade, or mask-based polygon crop.

\paragraph{Per-region follow-up threads.}
Each hotspot owns an isolated follow-up thread. A follow-up question is sent
to the text model with a \emph{local} context: the original question, the raw
answer, the clicked hotspot's detail, and that thread's history. The
answer is scoped to the selected region rather than to the entire image. Threads
are persisted alongside the ChatImage, enabling resumption across sessions.

\subsection{Implementation}
\label{sec:method_implementation}

\method{} is implemented as a model-agnostic web system. The browser
frontend is written in vanilla JavaScript; the local backend runs on Node.js,
proxies text, image, and vision services, and stores generated ChatImages,
hotspot metadata, alignment traces, and follow-up threads in SQLite. Vision
models are accessed through long-lived local adapters, which decouple browser
interaction from model checkpoints and service credentials. The reference
implementation supports both deterministic local execution for reproducible
tests and configured external services for full-generation runs. The code and
example artifacts are released with a project page for inspection and
reproduction.\footnote{\url{https://github.com/wencanjiang/ChatImage}}

\section{Experiments}
\label{sec:experiments}
\begingroup
\setlength{\textfloatsep}{8pt plus 2pt minus 2pt}
\setlength{\floatsep}{6pt plus 2pt minus 2pt}
\setlength{\intextsep}{6pt plus 2pt minus 2pt}

We evaluate \method{} along two axes. First, we measure interaction-loop
completion, defined as whether a benchmark prompt yields a complete ChatImage
consisting of a generated image and a hotspot layer. Second, we measure hotspot
evidence, defined as whether a clickable region is supported by an explicit
grounding source and usable preview geometry or mask data. Completion is
computed over valid full-generation runs; attempts terminated by upstream quota
limits are treated as external run-limit events rather than as failures of the
proposed interaction pipeline.

\subsection{Benchmark}
\label{sec:experiments_benchmark}

\begin{table}[H]
    \centering
    \scriptsize
    \setlength{\tabcolsep}{4pt}
    \renewcommand{\arraystretch}{1.07}
    \caption{\textbf{Benchmark composition and scenario coverage.} The
    30-question benchmark covers structured analytical answers
    (technical/business layouts), spatial guidance tasks, and object-rich
    everyday scenes.}
    \label{tab:benchmark_split}
    \begin{tabularx}{0.98\linewidth}{@{}lcc>{\raggedright\arraybackslash}X>{\raggedright\arraybackslash}X@{}}
        \toprule
        Mode & \#Q & Avg. hotspots & Scenario coverage & Alignment stress \\
        \midrule
        Infographic & 15 & 5.0 &
        technical workflows, comparisons, decision aids, checklists, timelines &
        compact panels, dense labels, semantically similar regions \\
        Map & 5 & 6.0 &
        tour routes, landmark clusters, spatial itineraries, scenic areas &
        route-like geometry, landmark placement, irregular spatial regions \\
        Scene & 10 & 6.0 &
        food options, coffee shops, reading spaces, record stores, plant care &
        object localization, overlapping items, natural scene composition \\
        \midrule
        Total & 30 & 5.5 & three visual modes & generation plus hotspot grounding \\
        \bottomrule
    \end{tabularx}
    \vspace{-0.6em}
\end{table}

The benchmark contains 30 questions designed to span distinct visual modes and
content types, as summarized in Table~\ref{tab:benchmark_split}. Map questions
stress geographic placement and route-like scenic regions; scene questions
stress object localization in less regular visual compositions; and
infographic questions stress compact analytical panels and decision-support
layouts. Each benchmark question is processed by the full \method{}
interaction pipeline using external text, image, and vision providers. The
public showcase is a curated subset produced by the same workflow and retained
only after strict visual-alignment screening.

Because the grounding step is model-agnostic
(Section~\ref{sec:method_alignment}), different regions can be aligned by
different signals: a dedicated localizer, a VLM grounding fallback, OCR, a
SAM-refined planned box, or a planned fallback. The evaluation reports both
completion and the source of each logged hotspot alignment. Representative
qualitative results are shown in
Figure~\ref{fig:qualitative}.

\begin{figure}[t!]
    \centering
    \includegraphics[width=1.0\textwidth]{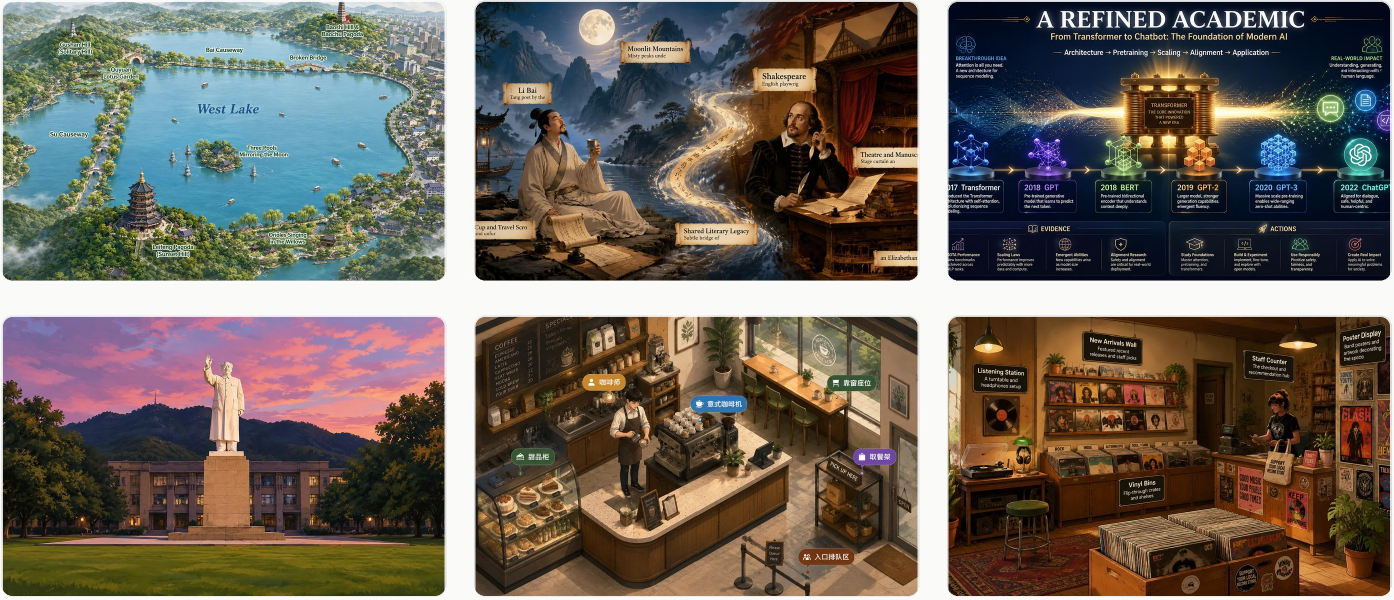}
    \caption{\textbf{Qualitative results.} Example \method{} outputs across
    diverse styles from the public demo gallery, including a hand-drawn map,
    literary comparison, academic poster, campus landmark scene, and everyday
    object-rich scenes. The examples use real generated demo images whose
    corresponding exported state supplies the interactive hotspot layer in the
    web interface.}
    \label{fig:qualitative}
    \vspace{-0.6em}
\end{figure}

\subsection{Interaction-Loop Results}
\label{sec:experiments_loop}

\begin{table}[H]
    \centering
    \footnotesize
    \setlength{\tabcolsep}{8pt}
    \renewcommand{\arraystretch}{1.05}
    \caption{\textbf{Hotspot quality diagnostic.} Rates use the same
    24-hotspot checked subset. "Strict-gate pass" counts hotspots accepted by
    the automatic grounding and mask-quality gate used before demo promotion.
    "SAM-complete" counts masks without holes or empty cavities. The
    complementary rows expose how much of the checked subset still needs better
    grounding or segmentation.}
    \label{tab:main_results}
    \begin{tabular}{@{}lccc@{}}
        \toprule
        Metric & Unit & Count & Rate \\
        \midrule
        \rowcolor{oursrowblue}
        \best{Strict-gate passes $\uparrow$} & Hotspot & \best{17/24} & \best{70.8\%} \\
        Strict-gate rejects $\downarrow$ & Hotspot & 7/24 & 29.2\% \\
        SAM-complete masks $\uparrow$ & Hotspot & 13/24 & 54.2\% \\
        SAM-incomplete masks $\downarrow$ & Hotspot & 11/24 & 45.8\% \\
        \bottomrule
    \end{tabular}
\end{table}

Figure~\ref{fig:experiment_summary} gives a compact visual summary of hotspot
quality, alignment-source mix, and the SAM mask-completeness diagnostic.

Across valid full-generation runs that reached the generation stage, all 30
benchmark questions produced a complete ChatImage with a generated image and
hotspot layer, yielding an interaction-loop completion rate of
\textbf{$100.0\%$}. Attempts interrupted by upstream quota limits are excluded
from this denominator. This result should be interpreted as a pipeline
completion diagnostic under valid provider responses; the more discriminative
criterion is hotspot reliability.

Table~\ref{tab:main_results} uses a shared 24-hotspot checked subset to report
both accepted and rejected regions under the strict visual-alignment gate. The
gate accepted \textbf{17 of 24 checked hotspots} ($70.8\%$) and rejected 7
($29.2\%$). The gate is intentionally conservative: an accepted hotspot must
have a primary grounding signal and usable preview geometry or mask data, so
weak or missing visual evidence is rejected rather than promoted. The
SAM-complete row reports a stricter mask-shape diagnostic on the same checked
subset. In total, \textbf{13} hotspots pass the SAM segmentation completeness
check ($54.2\%$), while 11 remain incomplete ($45.8\%$) due to obvious holes or
empty cavities.

\begin{figure}[t!]
    \centering
    \includegraphics[width=1.0\textwidth]{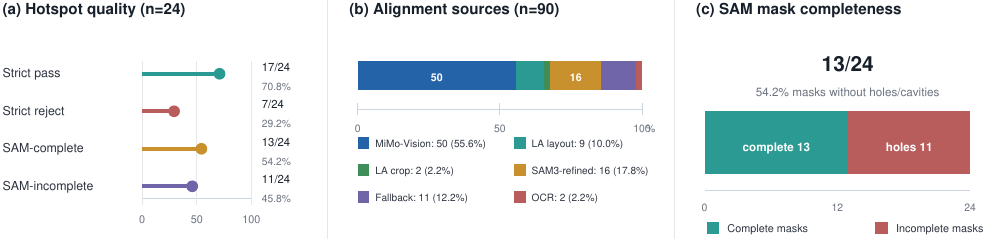}
    \caption{\textbf{Experimental metric summary.} Hotspot-quality diagnostics,
    alignment-source distribution, and the SAM mask-completeness check;
    SAM-complete requires masks without obvious holes or empty cavities.}
    \label{fig:experiment_summary}
    \vspace{-1.2em}
\end{figure}

\subsection{Alignment Source Analysis}
\label{sec:experiments_sources}

\begin{table}[H]
    \centering
    \footnotesize
    \setlength{\tabcolsep}{5pt}
    \renewcommand{\arraystretch}{1.05}
    \caption{\textbf{Alignment source distribution.} Source assigned to the
    90 logged hotspot alignments from the diagnostic source sample. Primary visual
    grounding combines MiMo-Vision and LocateAnything; planned-derived rows
    are retained by the robustness fallback when primary grounding is missing
    or rejected.}
    \label{tab:ablation_results}
    \begin{tabularx}{0.92\linewidth}{@{}Xcc@{}}
        \toprule
        Alignment source & Hotspots & Share \\
        \midrule
        MiMo-Vision primary grounding & 50 & 55.6\% \\
        LocateAnything layout-guided grounding & 9 & 10.0\% \\
        LocateAnything crop grounding & 2 & 2.2\% \\
        SAM3-refined planned region & 16 & 17.8\% \\
        Planned fallback & 11 & 12.2\% \\
        Local OCR support & 2 & 2.2\% \\
        \midrule
        \rowcolor{oursrowblue}
        \best{Total} & \best{90} & \best{\na} \\
        \bottomrule
    \end{tabularx}
\end{table}

Table~\ref{tab:ablation_results} reports the grounding-source distribution for
90 logged hotspot alignments from the diagnostic source sample. Most logged
hotspots are supplied by primary visual grounding: MiMo-Vision accounts for 50
hotspots ($55.6\%$), while LocateAnything contributes another 11 hotspots when
layout-guided and crop-based calls are combined ($12.2\%$). SAM3-refined
planned regions account for 16 hotspots ($17.8\%$), and planned fallback is
used for 11 hotspots ($12.2\%$). This distribution shows that \method{}
preserves the evidence used to place each hotspot: high-confidence visual
groundings are retained when available, plausible boxes are refined with masks,
and planned bounds are used only when stronger evidence is unavailable.

\FloatBarrier
\endgroup

\section{Conclusion}
\label{sec:conclusion}

\textbf{We introduced \method{}, a framework for transforming a long-form LLM
answer into an interactive visual image whose regions support localized
explanations and follow-up dialogue.} The central design principle is to avoid
treating pre-generation layout as final interaction geometry. Since an image
generator is not guaranteed to realize the planned layout faithfully,
\method{} first renders the answer and then grounds interaction regions in the
visible result. The implementation supports infographic, map, and scene modes,
and the evaluation reports interaction-loop completion, hotspot source
attribution, and a strict visual-alignment gate.

\noindent\textbf{Task positioning.}
\method{} frames interactive visual answering as an interaction and systems
problem rather than as a new generative-model architecture. Its contribution is
a compositional pattern: structure the answer, render it, ground the rendered
regions, and attach local interaction to each region.

\noindent\textbf{Use cases.}
The proposed format is most appropriate when an answer already has visual or
spatial structure. Educational explanations can expose stages of a process as
selectable regions; technical documentation can become a navigable artifact
with localized follow-up; spatial guidance can represent landmarks and routes
directly; and infographic answers can make comparisons scannable while
preserving drill-down detail. These use cases depend on the same assumption:
the visual artifact should be inspected as an interface, not only as an
illustration. Improvements in text-to-image generation and visual grounding are
therefore expected to directly improve the fidelity of \method{} outputs.

\noindent\textbf{Limitations and future work.}
The current system remains limited by the vision models on which alignment
depends. Dense infographics, small objects, and complex maps can still produce
weak or partial groundings. Click targets are also rectangular in the normal
interaction loop; polygon-level hit testing would better represent irregular
regions when masks are reliable enough to support interaction, not merely
preview. Finally, file context is primarily text-oriented, leaving richer
document parsing for future work. Promising directions include stronger
OCR-aware grounding, mask-based interaction, automated visual QA for hotspot
accuracy, shareable HTML exports, and synchronized cross-device history. The
open-source release and benchmark are intended to support inspection,
reproduction, and extension of this interaction format.

\begingroup
\renewcommand{\bibfont}{\scriptsize}
\setlength{\bibsep}{0pt}
\bibliography{chatimage}

@article{openai2023gpt4,
  title   = {{GPT-4} Technical Report},
  author  = {{OpenAI}},
  journal = {arXiv preprint arXiv:2303.08774},
  year    = {2023}
}

@article{anthropic2024claude,
  title   = {The {Claude} Model Family},
  author  = {{Anthropic}},
  journal = {Anthropic Technical Report},
  year    = {2024}
}

@inproceedings{wei2022cot,
  title     = {Chain-of-Thought Prompting Elicits Reasoning in Large Language Models},
  author    = {Wei, Jason and Wang, Xuezhi and Schuurmans, Dale and Bosma, Maarten and Ichter, Brian and Xia, Fei and Chi, Ed H and Le, Quoc V and Zhou, Denny},
  booktitle = {Advances in Neural Information Processing Systems},
  year      = {2022}
}

@inproceedings{yao2023react,
  title     = {{ReAct}: Synergizing Reasoning and Acting in Language Models},
  author    = {Yao, Shunyu and Zhao, Jeffrey and Yu, Dian and Du, Nan and Shafran, Izhak and Narasimhan, Karthik and Cao, Yuan},
  booktitle = {International Conference on Learning Representations},
  year      = {2023}
}

@inproceedings{schick2023toolformer,
  title     = {Toolformer: Language Models Can Teach Themselves to Use Tools},
  author    = {Schick, Timo and Dwivedi-Yu, Jane and Dess{\`\i}, Roberto and Raileanu, Roberta and Lomeli, Maria and Hambro, Eric and Zettlemoyer, Luke and Cancedda, Nicola and Scialom, Thomas},
  booktitle = {Advances in Neural Information Processing Systems},
  year      = {2023}
}

@article{jiang2026spike,
  title   = {{SPIKE}: An Adaptive Dual Controller Framework for Cost-Efficient Long-Horizon Game Agents},
  author  = {Jiang, Wencan and Zhang, Jiangning and Mei, Jianbiao and Liu, Jinzhuo and Yang, Yu and Hu, Xiaobin and Xue, Zhucun and Liu, Yong and Tao, Dacheng},
  journal = {arXiv preprint arXiv:2605.18636},
  year    = {2026}
}

@inproceedings{lewis2020rag,
  title     = {Retrieval-Augmented Generation for Knowledge-Intensive {NLP} Tasks},
  author    = {Lewis, Patrick and Perez, Ethan and Piktus, Aleksandra and Petroni, Fabio and Karpukhin, Vladimir and Goyal, Naman and K{\"u}ttler, Heinrich and Lewis, Mike and Yih, Wen-tau and Rockt{\"a}schel, Tim and Riedel, Sebastian and Kiela, Douwe},
  booktitle = {Advances in Neural Information Processing Systems},
  year      = {2020}
}

@article{gao2024ragsurvey,
  title   = {Retrieval-Augmented Generation for Large Language Models: A Survey},
  author  = {Gao, Yunfan and Xiong, Yun and Gao, Xinyu and Jia, Kangxiang and Pan, Jinliu and Bi, Yuxi and Dai, Yi and Sun, Jiawei and Wang, Meng and Wang, Haofen},
  journal = {arXiv preprint arXiv:2312.10997},
  year    = {2024}
}

@inproceedings{xie2026adakd,
  title     = {{LLM}-Oriented Token-Adaptive Knowledge Distillation},
  author    = {Xie, Xurong and Xue, Zhucun and Wu, Jiafu and Li, Jian and Wang, Yabiao and Hu, Xiaobin and Liu, Yong and Zhang, Jiangning},
  booktitle = {Proceedings of the AAAI Conference on Artificial Intelligence},
  year      = {2026},
  volume    = {40},
  pages     = {16676--16684},
  doi       = {10.1609/aaai.v40i16.40701}
}

@article{ramesh2022dalle2,
  title   = {Hierarchical Text-Conditional Image Generation with {CLIP} Latents},
  author  = {Ramesh, Aditya and Dhariwal, Prafulla and Nichol, Alex and Chu, Casey and Chen, Mark},
  journal = {arXiv preprint arXiv:2204.06125},
  year    = {2022}
}

@inproceedings{saharia2022imagen,
  title     = {Photorealistic Text-to-Image Diffusion Models with Deep Language Understanding},
  author    = {Saharia, Chitwan and Chan, William and Saxena, Saurabh and Li, Lala and Whang, Jay and Denton, Emily and Ghasemipour, Seyed Kamyar and Ayan, Burcu Karagol and Mahdavi, S Sara and Sarrafpour, R Konuk and others},
  booktitle = {Advances in Neural Information Processing Systems},
  year      = {2022}
}

@inproceedings{rombach2022ldm,
  title     = {High-Resolution Image Synthesis with Latent Diffusion Models},
  author    = {Rombach, Robin and Blattmann, Andreas and Lorenz, Dominik and Esser, Patrick and Ommer, Bj{\"o}rn},
  booktitle = {IEEE/CVF Conference on Computer Vision and Pattern Recognition},
  year      = {2022}
}

@inproceedings{podell2024sdxl,
  title     = {{SDXL}: Improving Latent Diffusion Models for High-Resolution Image Synthesis},
  author    = {Podell, Dustin and English, Zion and Lacey, Kyle and Blattmann, Andreas and Dockhorn, Tim and M{\"u}ller, Jonas and Penna, Joe and Rombach, Robin},
  booktitle = {International Conference on Learning Representations},
  year      = {2024}
}

@inproceedings{zhang2023controlnet,
  title     = {Adding Conditional Control to Text-to-Image Diffusion Models},
  author    = {Zhang, Lvmin and Rao, Anyi and Agrawala, Maneesh},
  booktitle = {IEEE/CVF International Conference on Computer Vision},
  year      = {2023}
}

@inproceedings{li2023gligen,
  title     = {{GLIGEN}: Open-Set Grounded Text-to-Image Generation},
  author    = {Li, Yuheng and Liu, Haotian and Wu, Qingyang and Mu, Fangzhou and Yang, Jianwei and Gao, Jianfeng and Li, Chunyuan and Lee, Yong Jae},
  booktitle = {IEEE/CVF Conference on Computer Vision and Pattern Recognition},
  year      = {2023}
}

@article{chen2024trainingfree,
  title   = {Training-Free Layout-to-Image Generation with Latent Diffusion Models},
  author  = {Chen, Weixuan and Liu, Bohan and Zeng, Daohan and Li, Shitian and Huang, Liangli and Xu, Chang and Huang, Chunjie},
  journal = {arXiv preprint arXiv:2401.13391},
  year    = {2024}
}

@article{cheng2024intelligent,
  title   = {Intelligent {IG}: Instruction-driven visual infographic generation and editing},
  author  = {Cheng, Yifu and Zhou, Peng and Cai, Jiafei and Wei, Zhongjie and Sun, Wei and Tan, Robby T and Hua, Gang},
  journal = {arXiv preprint arXiv:2312.14390},
  year    = {2024}
}

@inproceedings{feng2023layoutgpt,
  title     = {{LayoutGPT}: Compositional Visual Planning and Generation with Large Language Models},
  author    = {Feng, Weixi and Zhu, Wanrong and Fu, Tsu-Jui and Jampani, Varun and Akula, Arjun and He, Xuehai and Basu, Sugato and Wang, Xin Eric and Bernstein, Michael S},
  booktitle = {Advances in Neural Information Processing Systems},
  year      = {2023}
}

@article{lu2023infographicnlp,
  title   = {{InfographicNLP}: Expanding the Scope of Infographic Analysis},
  author  = {Lu, Simeng and Mancini, Michele and Zisserman, Andrew and Fritz, Mario},
  journal = {arXiv preprint arXiv:2310.13632},
  year    = {2023}
}

@inproceedings{kamath2021mdetr,
  title     = {{MDETR}: Modulated Detection for End-to-End Multi-Modal Understanding},
  author    = {Kamath, Aishwarya and Singh, Mannat and LeCun, Yann and Synnaeve, Gabriel and Misra, Ishan and Caron, Nicolas},
  booktitle = {IEEE/CVF International Conference on Computer Vision},
  year      = {2021}
}

@inproceedings{deng2021transvg,
  title     = {{TransVG}: Visual Grounding with Transformers},
  author    = {Deng, Zhengyuan and Wang, Jun and Zhang, Zihang and He, Yufan and Feng, Jianhua},
  booktitle = {IEEE/CVF International Conference on Computer Vision},
  year      = {2021}
}

@inproceedings{yan2023uninext,
  title     = {{UNINEXT}: Universal Instance Perception as Object-in-Context Prompting},
  author    = {Yan, Bin and Zhu, Yi and Wang, Jiangshan and Li, Zhihang and Zhang, Yichao and Liu, Wu and Zhang, Erjin and Sun, Jian},
  booktitle = {IEEE/CVF Conference on Computer Vision and Pattern Recognition},
  year      = {2023}
}

@inproceedings{liu2024groundingdino,
  title     = {Grounding {DINO}: Marrying {DINO} with Grounded Pre-Training for Open-Set Object Detection},
  author    = {Liu, Shilong and Zeng, Zhaoyang and Ren, Tianhe and Li, Feng and Zhang, Hao and Yang, Jie and Li, Chunyuan and Yang, Jianwei and Su, Hang and Zhu, Jun and Zhang, Lei},
  booktitle = {European Conference on Computer Vision},
  year      = {2024}
}

@article{cheng2024yoloworld,
  title   = {{YOLO-World}: Real-Time Open-Vocabulary Object Detection},
  author  = {Cheng, Tianheng and Song, Liming and Ge, Yixiao and Liu, Wenhao and Wang, Xinggang and Shan, Ying},
  journal = {IEEE/CVF Conference on Computer Vision and Pattern Recognition},
  year    = {2024}
}

@misc{nvidia2025locateanything,
  title        = {{LocateAnything-3B}: A Visual Grounding Model},
  author       = {{NVIDIA}},
  year         = {2025},
  howpublished = {Hugging Face model card, \url{https://huggingface.co/nvidia/LocateAnything-3B}},
  note         = {Accessed 2026}
}

@misc{xiaomi2025mimovl,
  title        = {{MiMo-VL}: Xiaomi {MiMo} Vision-Language Model},
  author       = {{Xiaomi LLM-Core Team}},
  year         = {2025},
  howpublished = {Model release, \url{https://github.com/XiaomiMiMo/MiMo-VL}},
  note         = {Used in this work via the \texttt{mimo-v2.5} chat-completion vision endpoint}
}

@inproceedings{liu2023llava,
  title     = {Visual Instruction Tuning},
  author    = {Liu, Haotian and Li, Chunyuan and Wu, Qingyang and Lee, Yong Jae},
  booktitle = {Advances in Neural Information Processing Systems},
  year      = {2023}
}

@inproceedings{chen2024internvl,
  title   = {{InternVL}: Scaling up Vision Foundation Models and Aligning for Generic Visual-Linguistic Tasks},
  author  = {Chen, Zhe and Wu, Jiannan and Wang, Wenhai and Su, Weijie and Chen, Guo and Xing, Sen and Zhong, Muyan and Zhang, Qinglong and Zhu, Xizhou and Lu, Lewei and others},
  booktitle = {IEEE/CVF Conference on Computer Vision and Pattern Recognition},
  year    = {2024}
}

@article{xu2026uniicl,
  title   = {{UniICL}: Systematizing Unified Multimodal In-context Learning through a Capability-Oriented Taxonomy},
  author  = {Xu, Yicheng and Zhang, Jiangning and Xue, Zhucun and Hu, Teng and Yi, Ran and Hu, Xiaobin and Liu, Yong and Tao, Dacheng},
  journal = {arXiv preprint arXiv:2603.24690},
  year    = {2026}
}

@inproceedings{kirillov2023sam,
  title     = {Segment Anything},
  author    = {Kirillov, Alexander and Mintun, Eric and Ravi, Nikhila and Mao, Hanzi and Rolland, Chloe and Gustafson, Laura and Xiao, Tete and Whitehead, Spencer and Berg, Alexander C. and Lo, Wan-Yen and Doll{\'a}r, Piotr and Girshick, Ross},
  booktitle = {IEEE/CVF International Conference on Computer Vision},
  year      = {2023}
}

@article{ravi2024sam2,
  title   = {{SAM 2}: Segment Anything in Images and Videos},
  author  = {Ravi, Nikhila and Gabeur, Valentin and Hu, Yuan-Ting and Hu, Ronghang and Ryali, Chaitanya and Ma, Tengyu and Khedr, Haitham and R{\"a}dle, Roman and Rolland, Chloe and Gustafson, Laura and others},
  journal = {arXiv preprint arXiv:2408.00714},
  year    = {2024}
}

@article{hochheiser2004interactive,
  title   = {Interactive Poster Visualization with {PhD Online}},
  author  = {Hochheiser, Harry and Shneiderman, Ben},
  journal = {IEEE Computer Graphics and Applications},
  year    = {2004}
}

@book{munzner2014visualization,
  title     = {Visualization Analysis and Design},
  author    = {Munzner, Tamara},
  publisher = {CRC Press},
  year      = {2014}
}

@inproceedings{antol2015vqa,
  title     = {{VQA}: Visual Question Answering},
  author    = {Antol, Stanislaw and Agrawal, Aishwarya and Lu, Jiasen and Mitchell, Margaret and Batra, Dhruv and Zitnick, C Lawrence and Parikh, Devi},
  booktitle = {IEEE/CVF International Conference on Computer Vision},
  year      = {2015}
}

@article{goyal2017vqav2,
  title   = {Making the {V} in {VQA} Matter: Elevating the Role of Image Understanding in {Visual Question Answering}},
  author  = {Goyal, Yash and Khot, Tejas and Summers-Stay, Douglas and Batra, Dhruv and Parikh, Devi},
  journal = {arXiv preprint arXiv:1612.00837},
  year    = {2017}
}

@inproceedings{mathew2021docvqa,
  title     = {{DocVQA}: A Dataset for {VQA} on Document Images},
  author    = {Mathew, Minesh and Karatzas, Dimosthenis and Jawahar, C.V.},
  booktitle = {IEEE/CVF Winter Conference on Applications of Computer Vision},
  year      = {2021}
}

@inproceedings{masry2022chartqa,
  title     = {{ChartQA}: A Benchmark for Question Answering about Charts with Visual and Logical Reasoning},
  author    = {Masry, Ahmed and Long, Do Xuan and Tan, Jia Qing and Joty, Shafiq and Hoque, Enamul},
  booktitle = {Findings of the Association for Computational Linguistics: ACL 2022},
  year      = {2022}
}

@article{zhang2014mindmap,
  title   = {Mindmap: A creative visual thinking tool for education},
  author  = {Zhang, Yong},
  journal = {Journal of Educational Technology Systems},
  year    = {2014}
}

@article{hu2024llmmindmap,
  title   = {Can Large Language Models Generate Visual Mind Maps from Text?},
  author  = {Hu, Yiyang and Pang, Liang and Lan, Yanyan and Xu, Jiafeng and Cheng, Xueqi},
  journal = {arXiv preprint arXiv:2402.10988},
  year    = {2024}
}

@inproceedings{obeid2020charttotext,
  title     = {{Chart-to-Text}: Generating Textual Descriptions of Charts},
  author    = {Obeid, Hoque and Hoque, Enamul},
  booktitle = {Findings of the Association for Computational Linguistics: EMNLP 2020},
  year      = {2020}
}

@article{tang2023chartlama,
  title   = {{ChartLlama}: A Multimodal {LLM} for Chart Understanding and Generation},
  author  = {Tang, Yifan and Zhang, Ao and Wang, Hengzhu and Li, Juan and Hu, Zhijiang},
  journal = {arXiv preprint arXiv:2311.16483},
  year    = {2023}
}
\endgroup

\end{document}